# A COMPARATIVE EVALUATION OF TWO ALGORITHMS OF DETECTION OF MASSES ON MAMMOGRAMS

Guillaume Kom[1,4]*, Alain Tiedeu[2], Martin Kom[2], John Ngundam[3]

[1]LAIA, IUT-FV, PO Box 134 Bandjoun, University of Dschang, Cameroon
[2]: LETS, GRETMAT, ENSP, Université de Yaoundé I, BP 8390, Yaoundé-Cameroun
[3]: ACL, ENSP, Université de Yaoundé I, BP 8390, Yaoundé-Cameroun
[4]: The Abdus Salam International Centre for Theoretical Physics, Trieste, Italy
*: Corresponding author

## ABSTRACT

*In this paper, we implement and carry out the comparison of two methods of computer-aided-detection of masses on mammograms. The two algorithms basically consist of 3 steps each: segmentation, binarization and noise suppression using different techniques for each step. A database of 60 images was used to compare the performance of the two algorithms in terms of general detection efficiency, conservation of size and shape of detected masses.*

## 1. INTRODUCTION

Early detection is the key to improving breast cancer prognosis. Consequently many countries have established screening programs. These programs yield large volumes of mammograms and shortage of radiologists makes their reading labour intensive, time consuming and often inaccurate. Retrospective studies have shown that in current breast cancer screenings between 10% and 25% of the tumours are missed by radiologists [1]. A Computer-Aided-Detection (CAD) system that prompts suspicious regions can draw the attention of the radiologist to a tumour he might otherwise have overlooked. A number of publications in the literature have therefore been devoted to developing algorithms for computerized detection of masses in mammograms. A few of them are described below.

D. Pasquale et al. [2] built a CAD system for mass characterization. This is mainly on a gradient-based segmentation algorithm and on the neural classification of several features computed on the segmented mass. The value AUC (An area under the ROC-curve) of 0.805 is achieved for the whole database, according to the correctly segmented masses [2]. Wei Jun et al. [3] investigated the use of a two-stage gradient field analysis to identify suspicious masses. Wei Jun et al. also developed in [4]a two-view information fusion method to improve the performance of our CAD system for mass detection. Paquerault et al. [5] develop a bilateral pairing technique to help reduce false-positives identified by a single-view computer-aided detection (CAD) system for breast masses. The algorithms of Padayachee et al. [6] were applied to 68 matched pairs of cranio-caudal and medio lateral-oblique mammograms to detect masses. The first matching algorithm used texture measures extracted from a grey-level co-occurrence matrix and a Euclidean distance similarity metric. The second algorithm used a grey-level co-occurrence matrix and a mutual information similarity metric. The latter algorithm also performed remarkably well with the matching of malignant masses. Naghdy et al. [7] outlined the shape of the region of interest (mass in mammograms) by using feature extraction capability of the wavelet transform followed by a novel recursive-enhancement morphology algorithm to detect the masses and then applied morphology-based segmentation algorithm to the enhanced image to separate the mass from the normal breast tissues. Mudigonda et al. [8] proposed a method for the detection of masses in mammographic images that uses Gaussian smoothing and subsampling operations as preprocessing steps. The mass portions are segmented by establishing intensity links from the central portions of masses into the surrounding areas.

Techniques described above, exhibit limits with regards either to the specificity of detected masses (stellate, spiculate, round…), or to the characteristics of extracted masses (size, shape, number…). Taking into account these drawbacks, we have developed, tested and compared two new algorithms for mass detection.

## 2. MATERIALS AND METHODS

### 2.1. Database

The database consists of cranio-caudal and medio-lateral images which contained sometimes one or several masses marked by experienced radiologists. Those mammograms were selected from patient files based on visual criteria. The mammograms were recorded with a kodak MXB film 100/NIF/30 × 40 cm screen /film combination using an analogic mammography equipment "GE seno BUCKY 8 × 2y s/DAF 66". Each image was digitized from a film using a high quality Acer 640 BT scanner operating at spatial resolution of 300 dpi × 300 dpi with 8 bits per pixel, with a sampling aperture of 85 μm in diameter and 12 pixels/mm sampling rate. Presently, all image processing and analysis are performed in the host computer. For images larger than 1800 × 1440 pixels, one can display either a region of interest (ROI) at full resolution or the entire image at reduced resolution. All the programs were written in matlab.

### 2.2. Mass detection algorithms

In this section, a description is given of each of the two mass detection algorithms.

#### 2.2.1. Description of the first algorithm

The detection of masses is performed in three steps. The masses are first segmented as follows. The contrast of the original image is enhanced through a linear transformation enhancement filter. The contrast-enhanced image (CEI) is subtracted from the original image (OI) yielding the segmented image (SI) on which the background is attenuated while masses are preserved.

The second step of this algorithm consists of applying a local thresholding to binarize the segmented image. In the last step of this algorithm, the binarized-image is smoothen with a high pass and a median filters to discard false positives (FPs) (that means regions labelled as masses and which are not true masses).

#### 2.2.1.1. Enhancement Filter

The simplicity of the linear transformation approach comes from the fact that a large class of filters can be implemented according to the changes of constant values $\alpha$ and $a$, chosen to obtain the enhanced image. This approach was successfully used by KOM et al. [9].

Let $a$ and $b$ be two positive real numbers so that:

$$m = a*\log(1+m*b) \tag{1}$$

where $m$ is the maximum value of grey level. Here we work with 255.

Then, given a value of $a$,

$$b = (\exp(m/a)-1)/m \tag{2}$$

Given a constant $\alpha$, pixels in original image $OI(i,j)$ which grey level values are rescaled to belong to the interval [0,1], are modified as follows to obtain the enhanced image $CEI(i,j)$.

If $OI(i, j) < \alpha$,  $CEI(i, j) = a * \log[1 + b * OI(i, j)]$ (3)

If $OI(i, j) > \alpha$,  $CEI(i, j) = [\exp[OI(i, j)/a] - 1]/b$ (4)

After many testing several values of $a$ and $\alpha$ chosen empirically according to the pixel values of the original image, we obtain good results with $a = 10000$ and $\alpha = 0,3$.

#### 2.2.1.2. Binarization method

For each $SI(i, j)$ in the breast area, a decision is made to classify it into a potential mass pixel or a normal pixel by the following rule:

If $SI(i, j) \geq TH(i, j)$ and $SI_{dif} \geq MvoisiP$ then $SI(i, j)$ belongs to the suspicious area else $SI(i, j)$ belongs to the normal area

$TH(i, j)$ is an adaptive threshold value computed by the formula :

$$TH(i, j) = MvoisiP + \gamma * SI_{dif} \text{ with } SI_{dif} = SI_{max}(i, j) - SI_{min}(i, j) \quad (5)$$

$MvoisiP$ is an average of pixels intensity in a small neighbourhood around the pixel $SI(i, j)$ and $\gamma$ is a thresholding bias coefficient which is chosen empirically

### 2.2.2. Description of the second algorithm

The second algorithm uses a histogram modification enhancement technique and a segmentation method based on minimisation of inertia sum [10]. A median filter is then applied on the segmented-image to filter out FP.

#### 2.2.2.1. Enhancement Filter

The enhanced-image EI (i,j) is in the form :
$$EI(i, j) = \lambda + \gamma(OI(i, j)) \quad (6)$$

Where $\lambda = \frac{OI(i,j)_{inf} \times OI(i,j)_{max} - OI(i,j)_{sup} \times OI(i,j)_{min}}{OI(i,j)_{max} - OI(i,j)_{min}}$ (7)

And

$\gamma = \frac{OI(i,j)_{sup} - OI(i,j)_{inf}}{OI(i,j)_{max} - OI(i,j)_{min}}$ (8)

where $[OI_{inf}, OI_{sup}]$ is the range of values of grey level available while $[OI_{min}, OI_{max}]$ is the range of grey level values in the image.

This filter is particularly suitable because it remains relatively efficient even with original images with much noise.

#### 2.2.2.2. Binarization

Let $\Omega = \{0,1,...,L-1\}$ be the set of grey level values present in the image and $P = \{C_1, C_2,...,C_{N_C}\}$ a partition of $\Omega$ into classes. The problem lies in optimising P by identifying characteristics with which to label the different classes.

Lets $k$ be the grey-level and $h(k)$ a component of the histogram vector and c a class counter. With this method, we chose as optimisation criteria, the minimisation of the inertia sum of different classes W (P).

$$W(P) = \sum_{n=1}^{N_C} \sum_{k \in C_n} h(k)(k - G(C_n))^2 \quad (9)$$

where $G(C_n) = \dfrac{\sum_{k \in C_n} k(h(k))}{\sum_{k \in C_n} h(k)}$ (10)

is the average value in a given class $C_n$.

The chosen threshold for binarization corresponds to the value of k for which W(P) is minimum.

## 3. RESULTS AND DISCUSSION

### 3.1. Visual example of the effect of the processing steps

For this study, we shrunk digital mammograms with the resolution of 343 _m by averaging 4 × 4 pixels into one pixel. A 3-mm object in an original mammogram occupies 36 pixels in a digitized image, with 84 _m resolution. After reducing the image size of a factors of four, the object will occupy about 8–9 pixels. An object with a size of 8 pixels is expected to be detectable by many computer algorithms. Therefore, the shrinking step is applicable for mass cases and saves computation time. In this research, we used 60 images on which masses had previously been identified (for those which contained masses) by a team of radiologists. They were used to evaluate and compare the two algorithms in terms of their ability to conserve the size and shape of detected masses and their general efficiency measured by the ROC-curves technique. The reference being the results from the team of radiologists.

Figure 1 and 2 give visual examples of the effect of the processing steps of the first and second algorithms on an image

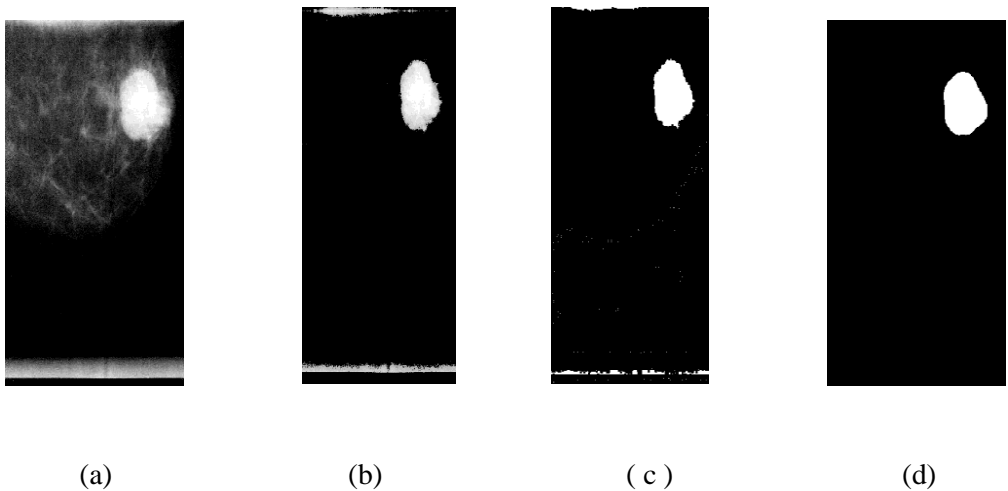

       (a)                (b)               ( c )              (d)

Figure 1. Visual example of the effect of the processing steps of the first algorithm. (a): original image, (b) : segmented-image, (c): binarized-image (d): mass detected

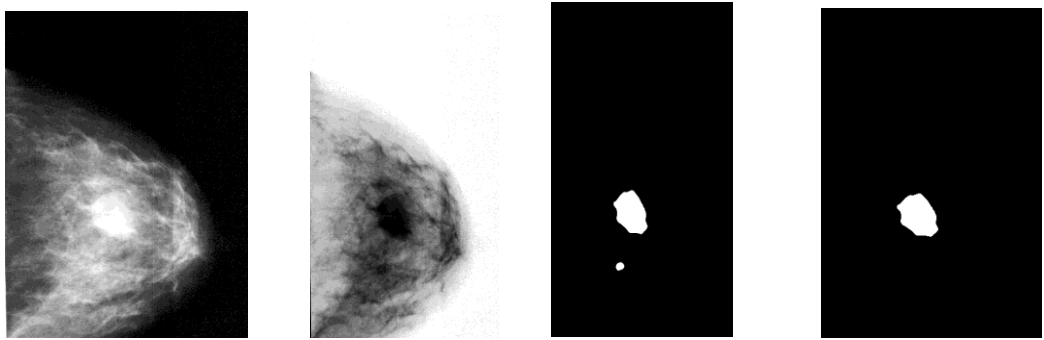

(a) (b) (c) (d)

Figure 2. Visual example of the effect of the processing steps of the first algorithm. (a): original image, (b) : enhanced-image, (c) : image after binarization, (d): mass detected.

### 3.2. Mass detection ability

The mammograms used in this research were obtained from two categories of breasts: very dense breasts and normal breast. We therefore tried to evaluate the efficiency of the two detection algorithms according to the breast density. **Table 1** recapitulates the detection ability of each algorithm according to the breast density.

Table 1. Performance of the two algorithms with respect to the breast density

|  |  | *Density of breast* | |
|---|---|---|---|
|  |  | Normal | Very dense |
| 1$^{st}$ Method | TP | 37 | 10 |
|  | FP | 2 | 0 |
|  | FN | 0 | 2 |
|  | TN | 20 | 7 |
|  | Sensitivity | 100 % | 83,3% |
|  | Specificity | 90,9% | 100% |
| 2$^{nd}$ Method | TP | 37 | 11 |
|  | FP | 2 | 1 |
|  | FN | 0 | 2 |
|  | TN | 20 | 7 |
|  | Sensitivity | 100% | 84,61% |
|  | Specificity | 90,9% | 87,5% |
|  |  |  |  |

For normal breasts, the two algorithms have the same performance. For very dense breasts, the first algorithm has a better sensitivity (capacity to detect masses where there are) and the second one a better specificity (capacity not to detect any masse where there is no mass).

It is difficult to conclude. We therefore used ROC (Receiver Operating Characteristics) evaluation to compare the overall performance of the two algorithms

Figure 3 summarizes detection performances of the two methods for the 60 images used. The standard masses are those detected by the radiologists.

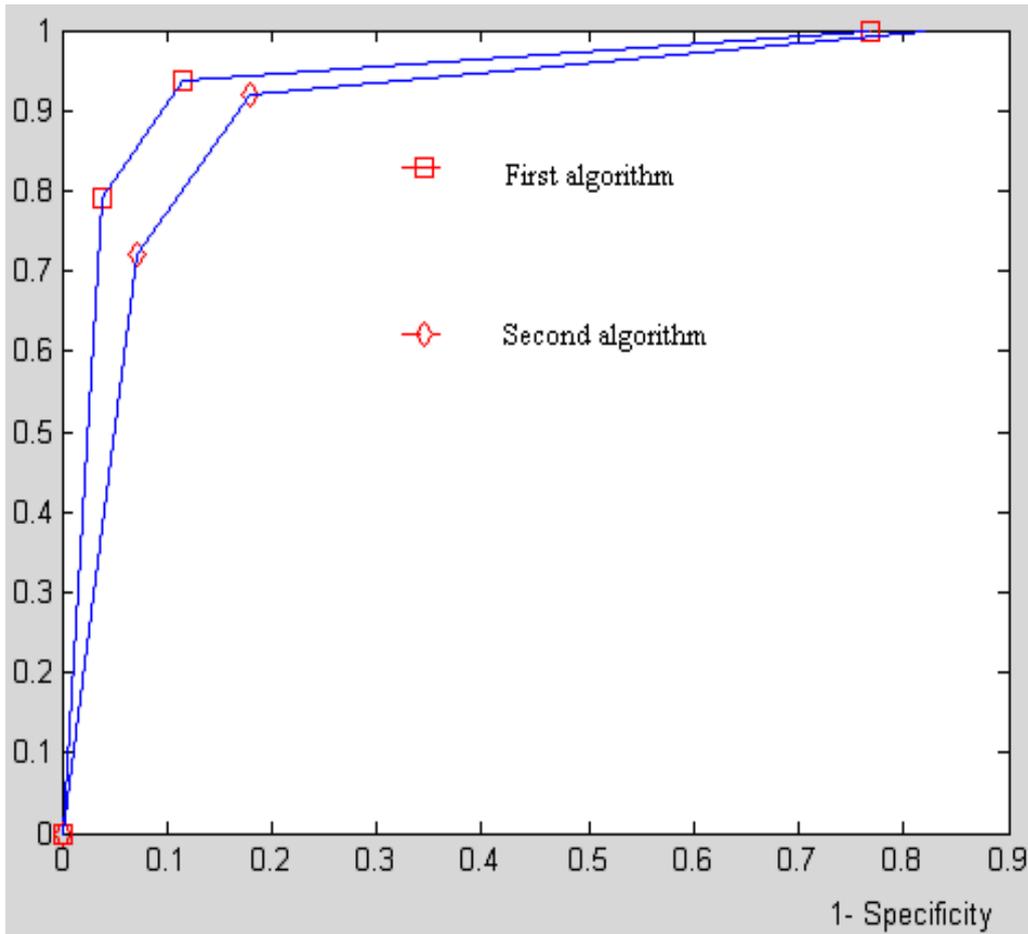

Figure 3. Comparison of mass detection performances for the two methods by ROC-curves

We noticed that the first method exhibits an area under the ROC-curve (0.94) greater than that of the second algorithm (0.91). May we suggest that this could be attributed to limited efficiency in fibrous regions of the breast. Overall, the first algorithm should be preferred in case there are a limited number of very dense breasts.

### 3.3 Conservation of the size and shape of masses detected

The two algorithms were also compared in terms of conservation of size and shape of masses detected. The standard sizes and shapes being given by the team of radiologists (**figures 4 and 5 below**) depict the results.

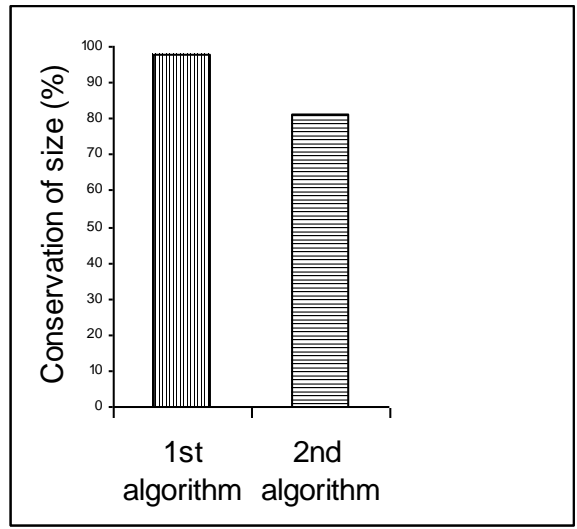

Figure 4. Comparison of ability to conserve the size of the detected masses by the two algorithms

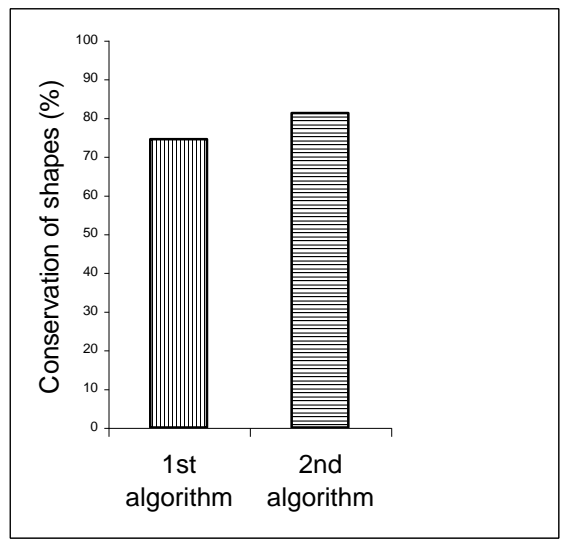

Figure 5. Comparison of ability to conserve shapes by the two algorithms (relative error on size < 5%)

The sizes of a good percentage (>80%) of mass detected are well conserved, which is a good point for both algorithms. However, for one quarter (25%) of detected masses, the shape is poorly rendered by the first algorithm. This is particularly due to masses with irregular edges like spicular masses. Knowing the importance of the shape and edges of masses for diagnosis, this is a serious drawback of this method which therefore suggests improvement of this algorithm. The possibility of combining the two methods is under consideration because, overall, the two algorithms look complementary.

## 4. CONCLUSION

Computer-aided-detection of masses in mammograms is emerging as one of the most promising approaches that may improve the efficacy of mammography. From drawbacks of methods in the literature, we have implemented and compared two detection algorithms. Although both give relatively good results in terms of detection of masses, conservation of shapes of masses still needs to be improved for the second algorithm. We anticipate that a combination of the two algorithms may give better results than that of each algorithm considered separately. We hope to investigate that in future research work.

## ACKNOWLEDGEMENTS

The authors wish to thank the Abdus Salam International Centre for Theoretical Physics (ICTP) and The Swedish International Development Agency (SIDA) for their financial support. They also thank Pr. JF GONSU and team of the HGOPY for providing the mammograms and the standard detection results.## REFERENCES

[1]   JA M van Dijck, L M Verbeek, Hendriks J H C L, and R Holland. "The Current detectability of breast cancer in a mammographic screening program". *Cancer*, 72:1933-1938, 1993.

[2]   D. Pasquale, E.F. Maria, K. Parnian, R. Alessandra. "Characterization of mammographic masses using a gradient-based segmentation algorithm and a neural classifier". *Computers in Biology and Medecine (CBM)*, vol.37, pp.1479- 491, 2007.

[3]   W. Jun, S. Berkman, H. Lubomir M., C. Heang-Ping, P. Nicholas, H. Mark A., Z. Chuan, G. Zhanyu. "Computer-aided detection of breast masses on full-field digital mammograms: false positive reduction using gradient field analysis". *Medical Imaging 2004: Image Processing, Edited by Fitzpatrick. J. Michael, Sonka. Milan. Proceedings of the SPIE*, Vol.5370, pp. 992-998, 2004.

[4]   W. Jun, S. Berkman, H. Lubomir M., C. Heang-Ping, H. Mark A., R. Marilyn A., Z. Chuan, G. Jun, Z. Yiheng. "Two-view information fusion for improvement of computer-aided detection (CAD) of breast masses on mammograms". *Medical Imaging 2006: Image Processing, Edited by Reinhardt. Joseph M., Pluim, Josien P. W. Proceedings of the SPIE,* Vol.6144, pp. 709-715, 2006.

[5]   S.Paquerault, N. Petrick, B. Sahiner, K. J. Myers, H.P. Chan."Potential improvement of computerized mass detection on mammograms using a bilateral pairing technique. *Medical Imaging 2006: Image Processing, Edited by Reinhardt, Joseph M., Pluim, Josien P. W. Proceedings of the SPIE*, Vol.6144, pp. 1821-1828, 2006.

[6]   J.Padayachee, M J Alport, W ID Rae. "Mammographic CAD: Correlation of regions in ipsilateral views - a pilot study". *South African Journal of Radiology*,Vol.13(3), 2009.

[7]   N.Golshah A., L. Yue, W. Jian."Wavelet-morphology for mass detection in digital mammogram images". *Medical Imaging 2003: Image Processing, Edited by Sonka. Milan, Fitzpatrick, J. Michael, Proceedings of the SPIE,*Vol. 5032, pp. 1313-1319, 2003.

[8]   N. R. Mudigonda, R.M. Rangayyan, J.E. Leo Desautels. "Detection of breast masses in mamograms by density slicing and texture flow-field analysis". *Medical Imaging, IEEE Transactions*, Vol.20 (12), pp: 1215–1227, 2001.

[9]   G. Kom, A Tiedeu, M. Kom.. "Automated detection of masses in mammograms by local adaptative tresholding" *Computers in Biology and Medicine* 37(1):37-48, 2007.

[10]   G. Kom, A. Tiedeu, M. Kom, C. Nguemgne, J. Gonsu. "Automated segmentation of masses in mammograms by minimisation of inertia sum". *ITBM-RBM* 26(.5-6):347-356, 2005.